\documentclass[manuscript,screen]{acmart}
\AtBeginDocument{%
  }

\setcopyright{acmlicensed}
\copyrightyear{2018}
\acmYear{2018}
\acmDOI{10.1145/xxxxxxx}
\acmJournal{TIOT}

\acmSubmissionID{123-A56-BU3}
\usepackage{threeparttable}
\usepackage{listings}  

\begin{document}

\title{A Hierarchical Test Platform for Vision Language Model (VLM)-Integrated Real-World Autonomous Driving}

\author{Yupeng Zhou}
\email{zhou1346@purdue.edu}
\orcid{0009-0000-3355-0955}
\affiliation{%
  \institution{Purdue University}
  \city{West Lafayette}
  \state{Indiana}
  \country{USA}
}
\author{Can Cui}
\email{cancui@purdue.edu}
\orcid{0009-0009-7082-3444}
\affiliation{%
  \institution{Purdue University}
  \city{West Lafayette}
  \state{Indiana}
  \country{USA}
}
\author{Juntong Peng}
\email{juntong@purdue.edu}
\orcid{0009-0007-1142-3067}
\affiliation{%
  \institution{Purdue University}
  \city{West Lafayette}
  \state{Indiana}
  \country{USA}
}
\author{Zichong Yang}
\email{zichong@purdue.edu}
\orcid{0009-0005-8019-3539}
\affiliation{%
  \institution{Purdue University}
  \city{West Lafayette}
  \state{Indiana}
  \country{USA}
}
\author{Juanwu Lu}
\email{juanwu@purdue.edu}
\orcid{0000-0003-0831-1244}
\affiliation{%
  \institution{Purdue University}
  \city{West Lafayette}
  \state{Indiana}
  \country{USA}
}
\author{Jitesh H Panchal}
\email{jpancha@purdue.edu}
\orcid{0000-0003-4873-3089}
\affiliation{%
  \institution{Purdue University}
  \city{West Lafayette}
  \state{Indiana}
  \country{USA}
}
\author{Bin Yao}
\email{byao@purdue.edu}
\orcid{0000-0003-3142-4570}
\affiliation{%
  \institution{Purdue University}
  \city{West Lafayette}
  \state{Indiana}
  \country{USA}
}
\author{Ziran Wang}
\email{ziran@ieee.org}
\orcid{0000-0003-2702-7150}
\affiliation{%
  \institution{Purdue University}
  \city{West Lafayette}
  \state{Indiana}
  \country{USA}
}


\renewcommand{\shortauthors}{Zhou et al.}

\begin{abstract}


Vision-Language Models (VLMs) have demonstrated notable promise in autonomous driving by offering the potential for multimodal reasoning through pretraining on extensive image–text pairs. However, adapting these models from broad web-scale data to the safety-critical context of driving presents a significant challenge, commonly referred to as domain shift. Existing simulation-based and dataset-driven evaluation methods, although valuable, often fail to capture the full complexity of real-world scenarios and cannot easily accommodate repeatable closed-loop testing with flexible scenario manipulation. In this paper, we introduce a hierarchical real-world test platform specifically designed to evaluate VLM-integrated autonomous driving systems. Our approach includes a modular, low-latency on-vehicle middleware that allows seamless incorporation of various VLMs, a clearly separated perception-planning-control architecture that can accommodate both VLM-based and conventional modules, and a configurable suite of real-world testing scenarios on a closed track that facilitates controlled yet authentic evaluations. We demonstrate the effectiveness of the proposed platform’s testing and evaluation ability with a case study involving a VLM-enabled autonomous vehicle, highlighting how our test framework supports robust experimentation under diverse conditions.

\end{abstract}





\maketitle

\section{Introduction}

Autonomous driving has emerged as a transformative technology with the potential to revolutionize transportation by enhancing safety, efficiency, and accessibility. In recent years, significant progress has been made in the development of autonomous driving systems, with advancements in perception, planning, and control algorithms. Deep learning techniques \cite{NIPS2012_c399862d} have greatly improved the accuracy and robustness of autonomous driving tasks, including object detection, semantic segmentation, and depth estimation.


Recent advances in Vision-Language Models (VLMs) are driving a paradigm shift in autonomous driving research \cite{Cui_2024_WACV}. Rather than training perception and policy models from scratch, the field is embracing a pre-train and then fine-tune paradigm. Foundation VLMs like CLIP \cite{radford2021learningtransferablevisualmodels} are pre-trained on hundreds of millions of image–text pairs, learning rich cross-modal representations and even demonstrating zero-shot generalization to new tasks. Researchers are now exploring ways to harness these powerful pre-trained models for autonomous vehicles \cite{10531702} because of such multimodal reasoning capabilities they offer for planning,  high-level semantic understanding and decision reasoning \cite{Pan_2024_CVPR, Park_2024_WACV}. The goal is to move from mere pattern recognition toward multimodal scene understanding and reasoning in the driving domain. VLMs bridge vision and language, enabling a model to comprehend complex relationships in visual scenes by leveraging aligned textual semantics \cite{Pan_2024_CVPR, Park_2024_WACV}.  More importantly, VLMs enabled an autonomous agent with a form of understanding closer to human-like reasoning, which is crucial for navigating complex, dynamic traffic scenarios.

However, realizing this vision requires a careful path from training to deployment: VLMs must be adapted from generic web-scale data to the safety-critical context of autonomous driving vehicles and rigorously tested under realistic conditions before they can be trusted on public roads, an issue which is often referred to as domain shift. The issue is highlighted by the discrepancy between the data distribution during training and that encountered during deployment, which poses a major challenge for VLM-integrated autonomous systems \cite{han2024anchorbasedrobustfinetuningvisionlanguage}. These discrepancies exist in various forms, such as variations in lighting conditions, weather patterns, road conditions, and object appearance. For VLMs, domain shift arises from two sources: The pre-training shift from general-purpose data to the driving world and the Fine-tuning shift from the broad pre-training distribution to a smaller, perhaps biased, driving dataset. During pre-training, VLMs learn from broad web-scale resources that cover diverse visual concepts but rarely the relevant autonomous driving messages and critical details of traffic scenes. The “open-world” imagery and language on the Internet differ significantly from the egocentric, safety-critical domain of driving. For example, an autonomous vehicle’s camera feed has specialized viewpoints (dashboard or roof-mounted cameras), motion blur, sensor noise, and a focus on road and traffic elements that generic datasets may not capture. During fine-tuning, the model is adapted to the driving domain, but this stage often relies on limited or synthetic data that introduces its own shifts. Researchers inject traffic-specific knowledge by fine-tuning domain datasets or even using VLM-generated annotations \cite{karnchanachari2024learningbasedplanningthenuplanbenchmark}, yet these fine-tuning sets are narrow in scope compared to the model’s original training diet. The net effect is that domain shift introduces uncertainty and blind spots in the VLM’s performance, undermining the very reliability that is paramount in autonomous driving. Bridging this gap requires careful domain adaptation techniques and, critically, extensive testing under conditions that mimic the target domain as closely as possible.

Evaluating VLM-integrated autonomous driving systems under controllable real-world conditions is, therefore, a non-trivial task because of the notable limitations of existing evaluation platforms. A majority of autonomous driving research relies on simulation environments such as Carla \cite{dosovitskiy2017carlaopenurbandriving}, AirSim \cite{shah2018airsim}, LGSVL \cite{9294422}. These simulation platforms provide configurable environments for testing and often lack the visual fidelity and behavioral complexity of real-world settings, leading to a noticeable sim-to-real gap. Models trained exclusively on synthetic data from these platforms tend to suffer degraded performance when faced with real-world data. As a complementary, many real-world datasets and benchmarks for autonomous driving assembled like the Waymo Open Dataset \cite{Sun_2020_CVPR}, Argoverse \cite{chang2019argoverse3dtrackingforecasting}, provide hours of recorded sensor data with annotations for tasks like object detection, tracking, and segmentation. Existing evaluation and testing platforms for VLM-integrated autonomous driving methods heavily rely on datasets and simulations during both the training and evaluation processes \cite{7795548, 9310544}. However, they are fundamentally logged data, lacking the ability to controllably reproduce interactions among traffic participants and autonomous driving agents or require them to adapt to changes in the evaluated autonomous driving agent’s behavior within a closed-loop setting. More recently, specialized benchmarks (e.g., NuScenes-QA \cite{Qian_Chen_Zhuo_Jiao_Jiang_2024} and nuPlan \cite{caesar2022nuplanclosedloopmlbasedplanning}) have started to incorporate multimodal and cognitive tasks. While these efforts mark progress, each covers only part of the autonomy stack. NuScenes-QA targets perception and situational reasoning but not vehicle control or navigation decisions; nuPlan enables the testing of planning algorithms but still within simulations. There remains a conspicuous gap in our evaluation methodology: we lack a comprehensive way to assess a vision-language-enabled autonomous vehicle as a whole under realistic real-world conditions with the ability to flexibly change scenarios and control experimental variables, which is crucial for effectively evaluating the impact of identified domain shifts and conducting targeted tests on specific domain shift scenarios.

In this paper, we propose a real-world hierarchical test platform for VLM-integrated autonomous driving to address the critical gap in testing and validation. Our platform is designed to test and validate VLM-based driving systems in controlled, authentic, and reproducible real-world scenarios prior to road deployment. The main contributions of the paper are:
\begin{itemize}
    \item We introduce a lightweight, structured, and low-latency middleware pipeline on the vehicle that is VLM-agnostic and modular, enabling seamless integration of various VLMs.
    \item Our architecture clearly separates the vehicle’s perception, planning, and control modules, allowing researchers to substitute VLM-based or conventional components at each stage.
    \item We develop a form of customizable real-world traffic scenarios on a closed test track and demonstrate the effectiveness of our framework through a case study featuring a real-world VLM-integrated autonomous vehicle.
\end{itemize}

The rest of this article is organized as follows: Section \ref{sec:Related Works} presents related works on state-of-the-art VLM-integrated autonomous driving based on VLMs, open-source autonomous driving stacks, and autonomous driving testing methods. Section \ref{sec:Methodology} describes the methodology, including the design of system architecture and testing environment. Section \ref{sec:Case Study} provides a detailed case study on the proposed testing framework deployment on each layer, including the real-world experiment setup. Section \ref{sec:Testing and Validation} covers testing and validation, evaluation metrics, and validation results. Finally, Section \ref{sec:Conclusion and Future Work} presents the conclusion, highlighting the key contributions and future work.

\section{Related Works}
\label{sec:Related Works}
\subsection{Autonomous Driving Stack}

An autonomous driving stack typically refers to the integrated software architecture that underpins an autonomous driving vehicle, encompassing essential modules such as perception, planning, and control. Because these functionalities are both complex and expansive, developing such a stack from scratch is time-consuming and costly. A more practical approach is to implement an open-source autonomous driving stack as a backbone, which offers a reliable foundation of core functionalities on top of which custom features can be developed. Open-source autonomous driving platforms have played a crucial role in advancing autonomous driving technology research and development by providing accessible, flexible, and collaborative frameworks for researchers and developers worldwide. These platforms can be broadly categorized into two main approaches: modular architectures and end-to-end systems \cite{chen2024end}. Modular architectures break down the autonomous driving stack into distinct components, allowing for targeted development and optimization of individual modules, while end-to-end systems aim to map sensory inputs to control outputs directly, leveraging the power of deep learning to create more integrated and streamlined pipelines. The development of autonomous driving vehicles relies heavily on these advanced software platforms that enable the integration of various perception, planning, and control algorithms, accelerating research and innovation in the field.

Modular architectures, such as Autoware \cite{autowarefoundation} and Baidu's Apollo \cite{apollo_ros}, provide a comprehensive framework that includes perception, planning, and control modules. Built on ROS and ROS2 (Robotic Operation System) \cite{quigley2009ros}, Autoware has gained significant adoption in academic and industrial settings due to its flexibility, extensibility, and strong community support. Similarly, Apollo offers a more commercially oriented approach, providing not only software frameworks but also hardware reference designs and simulation environments. These modular platforms allow researchers and developers to focus on specific components of the autonomous driving stack, enabling rapid innovation and collaboration.

On the other hand, end-to-end systems aim to map sensory inputs to control outputs directly, bypassing the need for explicit intermediate representations. Comma.ai's openpilot \cite{openpilot} exemplifies this approach, focusing specifically on advanced driver assistance systems (ADAS) for highway driving. Openpilot offers end-to-end, road-proven lateral and longitudinal control capabilities across multiple vehicle platforms, demonstrating the potential of end-to-end learning in real-world autonomous driving applications.

In addition to these comprehensive platforms, the field also includes several specialized frameworks targeting specific use cases. Roborace \cite{mar2024review} addresses autonomous racing at full scale, pushing the boundaries of high-speed autonomous navigation and control. F1TENTH \cite{okelly2020f1tenth} provides a 1:10 scale autonomous racing platform particularly suited for education and rapid algorithm prototyping, enabling students and researchers to experiment with autonomous driving algorithms in a controlled environment. Other notable platforms include Stanford's PASTA (Platform for Autonomous Systems and Testing Applications), which offers a modular framework specifically designed for testing various autonomous navigation algorithms \cite{8443742}. These diverse platforms, ranging from educational tools to production-ready frameworks, have collectively accelerated innovation in the autonomous driving domain. By providing researchers and developers with robust foundations for testing and implementing new algorithms, these platforms have facilitated the development of cutting-edge autonomous driving techniques.

\subsection{Autonomous Driving Testing}

As autonomous driving technology advances rapidly, it is crucial that autonomous driving systems are comprehensively tested to ensure they consistently behave correctly and safely, especially in edge cases where pedestrians and vehicles act unpredictably. Before autonomous vehicles can be fully deployed on public roads, rigorous testing is essential. The state-of-the-art in autonomous driving testing can be broadly categorized into two approaches: simulation-based testing and real-world testing and \cite{10064002}.

Simulation-based testing using high-fidelity platforms, such as CARLA \cite{dosovitskiy2017carlaopenurbandriving} and LGSVL \cite{9294422}, has gained popularity among researchers and developers. These simulations can replicate not only vehicle and sensor dynamics but also traffic conditions, allowing a wide range of testing research focusing on safety verification \cite{8569950}, communication \cite{gabbar2022modeling}, computation models \cite{9406805} and the generation of testing scenarios \cite{8500406}. Some testing frameworks are also designed for more specific tasks, such as planning \cite{o2016apex} and control \cite{li2019temporal}. Simulation allows for cost-effective testing of potentially dangerous scenarios, like accidents, without real-world risks. To further increase fidelity, Hardware-in-the-Loop (HIL) platforms \cite{8500461} are widely used for testing and validation.

On the other hand, extensive and rigorous real-world road testing is necessary for autonomous vehicles. It often serves as a measure of the technology's progress, such as the number of kilometers covered without human intervention. EDGAR \cite{karle2023edgar}, an autonomous driving platform for research, demonstrates comprehensive functionality in real-world deployment, testing, and validation for state-of-the-art autonomous driving technology.

Recent research on VLMs in autonomous driving spans a wide range of tasks and applications, from perception \cite{Liu_2023_CVPR, 10.1145/3664647.3681560, cheng2023languageguided3dobjectdetection, wu2023languagepromptautonomousdriving} and understanding to navigation \cite{10160614, omama2023altpilotautonomousnavigationlanguage, tian2024drivevlmconvergenceautonomousdriving}, decision-making \cite{9157238, sha2023languagempclargelanguagemodels}, motion control \cite{cui2024onboardvisionlanguagemodelspersonalized}, end-to-end learning \cite{10629039, wang2023drivemlmaligningmultimodallarge, Pan_2024_CVPR, xu2024vlmadendtoendautonomousdriving}, and data generation \cite{10531702}. Among these works, most are tested and validated using real-world datasets. Only a few works extend to on-board deployment and conduct real-world experiments \cite{tian2024drivevlmconvergenceautonomousdriving, 2023arXiv231209397C, cui2024onboardvisionlanguagemodelspersonalized}. As VLM-based autonomous driving technology rapidly grows, there is an urgent need for real-world testing and validation. In this work, we address this gap by extending an existing open-source autonomous driving stack into a hierarchical testing framework that enables experimentation and evaluation of VLM-based autonomous driving algorithms in real-world conditions. Our proposed framework bridges the gap between cutting-edge VLM-integrated autonomous driving research and practical deployment, providing a flexible and robust testing solution for researchers and developers to easily integrate, test, and evaluate these advanced techniques in repeatable real-world driving scenarios by leveraging the capabilities of an established open-source stack and augmenting it with VLM integration.
\begin{figure}[H]
    \centering
    \includegraphics[width=0.85\textwidth]{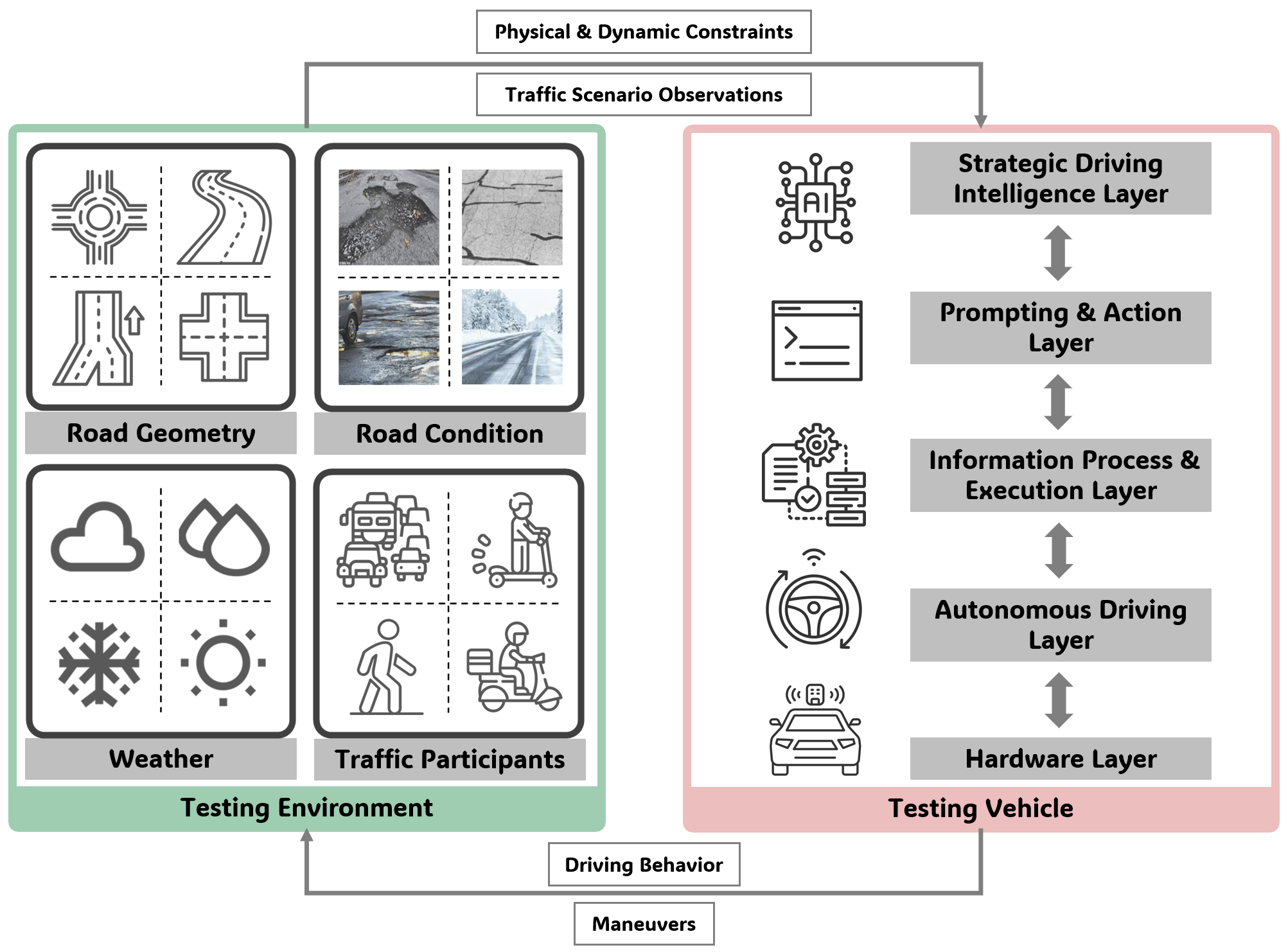}
    \caption{Overview of the proposed real-world testing platform for VLM-integrated autonomous driving comprising two main components: the testing environment and the testing vehicle.}
    \Description{}
    \label{fig:platform_overview}
\end{figure}
\section{Methodology}
\label{sec:Methodology}

Our real-world testing platform for VLM–integrated autonomous driving consists of two main components: the testing environment, and testing vehicles, as illustrated in Fig. \ref{fig:platform_overview}. The testing environment comprises road geometry, road conditions, weather, and traffic participants to generate diverse testing scenarios for evaluation and validation, providing both sensor observations and physical constraints that affect vehicle handling. By modifying these elements (e.g., introducing specific traffic patterns or altering visibility), one can systematically create or replicate diverse scenarios for more comprehensive evaluation and validation of the VLM agents. This controllability ensures that researchers can target particular driving conditions or repeat situations of interest to assess the under-testing VLM-integrated autonomous driving performance thoroughly.

\begin{figure}[H]
  \centering
  \includegraphics[width=\linewidth]{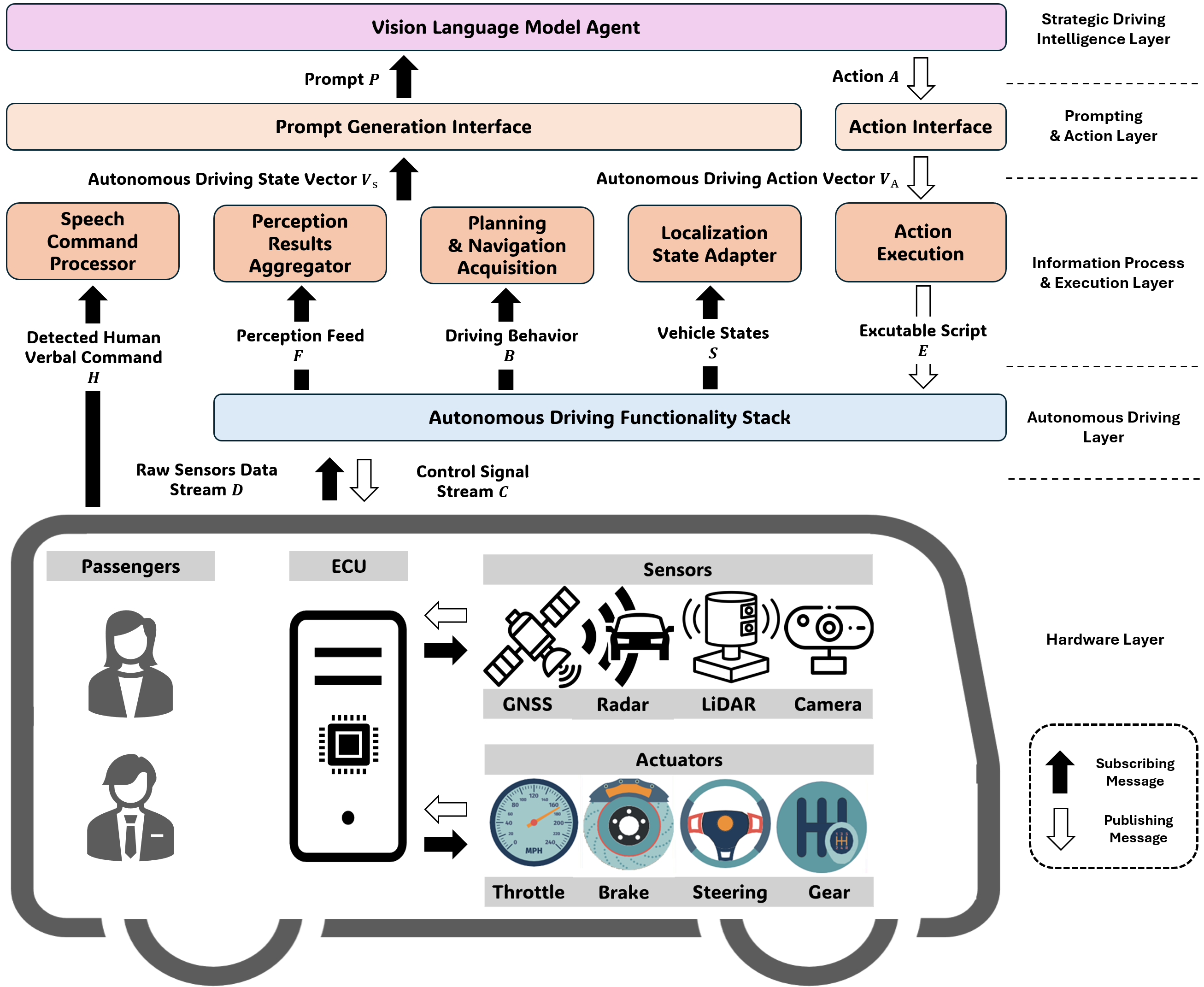}
  \caption{An overview of the proposed hierarchical architecture for VLM-integrated autonomous driving in the testing vehicle. The top-level VLM agent generates actions through the Prompting \& Action Layer, relying on real-time driving information provided by the Information Process \& Execution Layer. The Autonomous Driving Layer processes sensor data from the Hardware Layer including perception outputs, planned trajectories, and vehicle states—and executes the VLM’s commands to control the vehicle’s actuators, thus closing the loop between VLM-driven decisions and physical driving maneuvers.}
  \Description{Hierarchical Autonomous driving framework.}
  \label{fig:framework}
\end{figure}

From the perspective of the test vehicle, the hierarchical design balances the flexibility provided by the multilayer pipeline with the stability ensured by the classical autonomy components. By compressing both perceptual data captured by sensor kits in the hardware layer and environmental factors (e.g., traffic scenarios, visibility) recognized from the Autonomous Driving Layer, these inputs are processed through the hierarchical stack of the test vehicle: from the information process and execution layer to the VLM in strategic driving intelligence for high-level decision making. In return, the generated action will be executed by the low-level control from the hardware and autonomous driving layers to achieve the desired vehicle’s maneuvers and behaviors in the testing environment. This platform provides a robust closed-loop evaluation of VLM-integrated autonomous driving.

\subsection{Tesing Vehicle Architecture}

In this work, we present a hierarchical autonomous driving testing framework as shown in Fig.~\ref{fig:framework} that integrates a VLM with a classical autonomy pipeline for real-world, by-wire-enabled vehicles. The proposed framework operates through a series of interconnected processes that enable the system to reason about contextual and vision inputs of driving scenarios and execute appropriate maneuvers $\boldsymbol{C}$. The workflow begins with the raw sensory data $\boldsymbol{D}$ generated by the Hardware Layer. In the input pipeline, the input message processor will aggregate and abstract relevant driving information into a driving state vector $\boldsymbol{V_s}$, which includes perception feed $\boldsymbol{F}$, driving behavior $\boldsymbol{B}$, vehicle states $\boldsymbol{S}$ and along with human verbal command $\boldsymbol{H}$. Storing those driving information into a vector would not only make it easy for Prompting Interface to generate the prompt $\boldsymbol{P}$ but also reduce unnecessary content for efficiency. The VLM then interprets or reasons over $\boldsymbol{P}$ for tasks ranging from semantic scene understanding and route planning to conversing with onboard passengers or making real-time strategy decisions. The VLM outputs its reasoning results into an action vector $\boldsymbol{V_a}$, which captures high-level directives or recommendations. In the output pipeline, the action execution module reads these directives from $\boldsymbol{V_a}$ to produce an executable script $\boldsymbol{E}$. This script is sent to the Autonomous Driving Layer, where either a modular or end-to-end autonomy stack can incorporate the VLM’s guidance and handle fine-grained motion control or other operational tasks. Ultimately, the desired maneuvers $\boldsymbol{C}$ reach the vehicle actuators—closing the loop between VLM-driven decision-making and real-world driving behaviors and facilitating thorough evaluations across a spectrum of test scenarios.


\subsubsection{Strategic Driving Intelligence Layer:}

This layer is essentially the VLM itself, representing the top-level intelligence for high-level decision-making in the autonomous driving stack. In terms of composition, it can host either an on-board or on-cloud VLMs, allowing for flexible deployment and experimentation. The primary input is the structured prompt $\boldsymbol{P}$, which integrates multi-modal information such as perception results, vehicle states, driving behavior, and human feedback. The output is a high-level action decision $\boldsymbol{A}$ that is subsequently encoded into an action vector $\boldsymbol{V_A}$. The prompt $\boldsymbol{P}$ integrates multi-modal inputs, including vision-based inputs like perception results and contextual inputs that contain real-time driving behaviors, vehicle status, and human feedback. During testing, this layer serves to enable the real-time interaction of either on-board or on-cloud VLMs. 
    
By processing this rich context, the VLM determines an appropriate driving strategy and generates actions to adapt to complex, dynamic traffic conditions. Its primary function is strategic inference and planning, effectively offloading high-level reasoning from lower-level modules, which can continue to handle stable perception and control tasks. At the same time, this separated-layer design architecture is highly customizable; researchers can replace the VLM with different models to evaluate various reasoning capabilities, and the prompt/action interfaces (both input and output ports) are easily adjustable for VLMs with different requirements.

\subsubsection{Prompting \& Action Interface Layer:}

Acting as the intermediary between the Strategic Driving Intelligence Layer and the mid-level processes, this layer comprises two modules that operate together in a single workflow. In the input pipeline, the Prompting Interface packages and constructs structured prompts containing relevant information about the autonomous vehicles's state and environment. These prompts are passed to the VLM to elicit high-level decisions and plans. In the output pipeline, the Action Interface converts the action commands returned by the VLM into an autonomous driving action vector $\boldsymbol{V_A}$, which includes concrete parameters and instructions that downstream modules can execute. This layer effectively translates the abstract reasoning of the VLM into the precise inputs and outputs of the autonomous system.
    \begin{equation}
    \begin{aligned}
        \text{Promt Generation}:\quad & V_s \xleftarrow{Prompt- Interface} f_{process}(H,F,B,S); \\
        \text{Action Generation}:\quad & V_a \xleftarrow{Action-Interface} A \xleftarrow{\text{VLM}} P(V_s) ; \\
    \end{aligned}
    \end{equation}
These two phases above indicate how the Information Process \& Execution Layer work as a bidirectional bridge between the high-level VLM in the Driving Intelligence Layer and the mid-level process and action modules. During the actual tests, users can tune this layer independently to see how different prompt $\boldsymbol{P}$ would have affected the VLM behaviors with controller variables. This design also provides significant flexibility: for different VLMs, both the prompt-query structure and the output-action format can be adjusted to optimize integration and handle diverse testing scenarios and goals, and relevant messages can be dynamically added or removed. Such flexibility is essential for exploring how variations in prompt design and action representation affect VLM behavior and overall system performance.


\subsubsection{Information Process \& Execution Layer:} The Information Process \& Execution Layer implements a two-phase approach to interpret and act upon the high-level directives from the VLM. In the information processing phase, dedicated modules transform necessary messages and autonomous driving statistics into a vector named Autonomous Driving State Vector $\boldsymbol{V_s}$. These structured representations can be further processed by Prompt Generation Interface to integrate and inform the VLM's reasoning. In this input pipeline, our framework provides even more flexibility to revise the testing platform to specific desired testing goals. In particular, we provide comprehensive sub-processors aiming to grab pivot messages from autonomous driving function modules from the perception module and planning module to the control module.
Additionally, VLMs have revealed great potential in personalized autonomous driving by understanding the on-board passengers' verbal commands \cite{cui2024onboardvisionlanguagemodelspersonalized}. Thus, a Speech Command Processor is integrated into the Information Process Layer to capture the detected verbal commands $\boldsymbol{H}$ from humans. This layer dynamically collects and abstracts critical messages and information at a relatively low frequency from the autonomous driving functionality stack. These messages and data are then reformatted for easy integration into a well-formed prompt. During the real-world scenarios testing, the perception feeds $\boldsymbol{F}$ can be multiple forms of perception results,s including object trackers, raw image data, BEV image, and even the bounding boxes of various objects. The Driving Behavior $\boldsymbol{B}$ could be planned trajectories. The Vehicle States $\boldsymbol{S}$ could consist of vehicle speed, position, and even the map information. Researchers are free to form various combinations of these messages to a delicate Autonomous Driving State Vector $\boldsymbol{V_s}$, which is most suitable for the designed testing scenarios.
    
The execution phase converts the autonomous driving action vector $\boldsymbol{V_A}$  from the VLM into executable maneuvers by the Action Execution module. This module is composed of executable scripts or programs. Developers can leverage the available interfaces or ports in the lower Autonomous Driving Functionality Stack to design or develop scripts or programs tailored to specific test scenarios. These scripts or programs modify the inputs, outputs, or parameters of certain sub-functions within the perception, planning, and control modules in Autonomous Driving Layers, thus achieving the desired vehicular maneuvers as determined by the VLM. For example, during a real-world test, if the VLM outputs an action that includes a lane-change and overtaking command with selected feasible trajectories, the Driving Behavior Selection sub-module within the Action Execution will match the planning outcome accordingly. It then updates the selected correct trajectory with the control module inputs and adjusts related parameters, such as speed for the lane-change maneuver.

In our proposed testing platform, the execution phase interacts with the interfaces and ports exposed by the perception, planning, and control modules to ensure that the diverse commands generated by the VLM are effectively executed. Furthermore, the Action Execution module optimizes and validates the VLM's decisions against safety constraints before engaging the motion control layer. In real-world testing, these actions range from providing supplementary perception inputs and selecting driving behaviors to refining motion control. Fundamentally, to ensure safety, the Action Execution module also optimizes and validates the VLM's decisions against safety constraints before engaging motion control in the next layer.

\subsubsection{Autonomous Driving Layer:} 
The Autonomous Driving Layer contains the core autonomous driving functionality, which can be implemented using a classical module-based approach. In a module-based system, the Autonomous Driving Layer includes distinct subsystems for perception, planning, and control to provide comprehensive essential autonomous driving ability, including object detection, trajectory planning, motion control, and so on. It ingests raw sensor data or intermediate perception results, interprets the surrounding environment, and computes feasible trajectories based on well-defined rules and constraints. Alternatively, an end-to-end autonomous driving stack can be employed in the Autonomous Driving Layer. However, due to the high costs and substantial workload required to redevelop a complete and comprehensive autonomous software stack, our proposed testing platform utilizes an open-source autonomous driving platform as its backbone. This platform provides a reliable foundation of core functionalities on which custom features can be built.

To facilitate the execution of commands and code from the upper Action Execution Layer by the lower autonomous driving functionalities, we develop a series of interfaces and ports for each of the perception, planning, and control modules that perform specific autonomous driving tasks. These interfaces expose various parameters, input data, and output results. In real-world experiments, these elements may include sensor fusion parameters, filter settings, detection and classification thresholds, bounding boxes, and object trackers from the perception module; parameters for path smoothness, turning radius, and feasible trajectories from the planning module, as well as controller parameters and constraints from the control module. The wide range of adjustable parameters, input data, and output results, along with the plug-in ability for each sub-module, offers developers scalable flexibility to select, create, and fine-tune settings tailored to their specific testing objectives. 

\subsubsection{Hardware Layer:}
Finally, at the lowest level, the Hardware Layer encompasses the physical sensor kits, computing platforms, and actuators of the vehicle. Those sensor kits usually consist of LiDAR, radar, and cameras, capture raw sensor data, which is passed upward to the Autonomous Driving Layer to generate the control signal to follow the final trajectory, which is realized through actuation signals (e.g., steering, throttle, braking) sent to the vehicle's drive-by-wire system. By isolating hardware‐specific concerns, the framework can be readily adapted across diverse vehicle platforms and sensor configurations.

In summary, given the multi-layered architecture, it is crucial to rigorously evaluate the latency of message transmission across each layer to ensure that communication overhead does not adversely affect system performance. In parallel, the evaluation must verify that the system’s responses accurately capture and reflect user requirements. The detailed experiments for this evaluation will be presented in Section \ref{sec:Testing and Validation}.

\subsection{Testing Enviroment Setup}
Our scenario-based testing methodology systematically constructs diverse real-world conditions by varying four key parameters: road geometry, road conditions, weather, and traffic participants. This approach allows for both routine and edge-case scenarios to be tested under realistic conditions. We consider common driving maneuvers such as lane changes, car-following, and turns at intersections, in addition to more specialized configurations like parking lots, multi-lane highways, and complex road merges. More advanced, we can safely stage a critical near-crash scenario with dummy pedestrians or remote-controlled vehicles multiple times to see if a VLM-based system consistently makes the correct decision. These deployments in the physical world are something neither open-road testing nor conventional simulations easily allow. On the other hand, by adjusting road conditions (e.g., potholes, wet or uneven surfaces) and road geometry (e.g., sharp turns, narrow lanes, steep grades), we can precisely control the level of challenge presented to the vehicle’s perception, planning, and control systems.

To capture realistic environmental variability, weather conditions such as rain, snow, and fog are introduced, affecting sensor performance (e.g., camera occlusion, degraded LiDAR returns) and vehicle dynamics (e.g., reduced traction, longer braking distances). Moreover, traffic participants, ranging from pedestrians and cyclists to scooters and various classes of vehicles, can be flexibly added or removed to increase traffic complexity and test interaction behaviors under constrained road spaces. These dynamic elements are essential for thoroughly evaluating safety, particularly in situations involving multiple road users with potentially unpredictable trajectories.

Through systematic iteration of test runs across these scenario configurations, we can ensure repeatability for direct performance comparisons while retaining precise control over the variables involved. This methodical approach makes it possible to rigorously evaluate the performance and reliability of the under-testing VLM within consistent conditions, thereby isolating the impact of each parameter (e.g., traffic density, road surface condition, visibility) on overall system performance. By maintaining strict control of these environmental factors, we obtain reliable, reproducible insights that would be challenging to achieve through open-road testing or purely synthetic simulations.

In practice, each parameter can be systematically tuned to address different operational design domains and domain shift challenges. For instance, low-visibility weather not only tests sensor robustness but also pushes the vehicle’s decision-making to operate with degraded information. Likewise, introducing diverse traffic behaviors ensures the vehicle’s strategic driving intelligence handles a broad spectrum of social interactions. Through parametric variation and careful combination of these factors, we achieve a highly configurable testing framework, which generates scenarios ranging from basic car-following on well-maintained roads to extreme corner cases involving sharp curves, sudden cut-ins, or significant weather disruptions. This breadth of scenarios provides a comprehensive validation ground for VLM-integrated autonomous driving systems, ensuring that both perception-driven and language-based reasoning modules are rigorously challenged.

\begin{figure}[H]
  \centering
  \includegraphics[width= \linewidth]{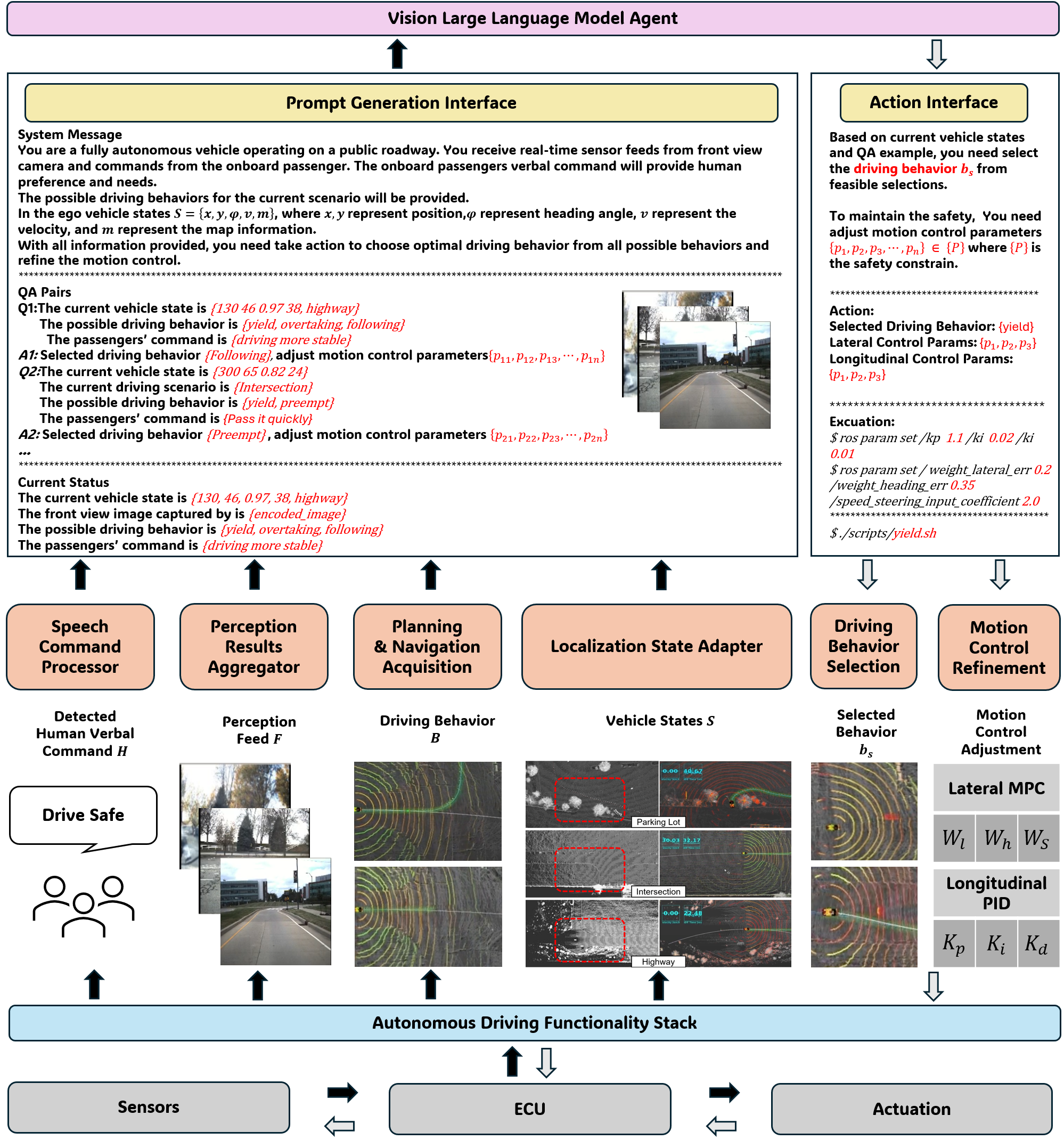}
  \caption{A detailed illustration of our proposed hierarchical autonomous driving testing frame is applied to those intermediate layers in the case study to verify how the VLM makes driving decisions under complex scenarios and handles the vehicle motion control tasks, as an example.}
  \Description{In this example, the Information Process Layer aggregates perception outputs,safety-guaranteed trajectories generated by the planning module, vehicle states, and human feedback, then formats them into a structured prompt via the Prompt Generation Interface. The top-level VLM uses this prompt to generate driving decisions, which the Action Interface captures and forwards to the Driving Behavior Selection and Motion Control Refinement modules. These modules invoke the Autonomous Driving Functionality Stack, including planning, localization, and actuation, to execute the desired maneuvers on the physical vehicle.}
  \label{fig:case}
\end{figure}

\section{Case Study}
\label{sec:Case Study}
In this paper, to illustrate the performance and effectiveness of the proposed testing platform, we implement the proposed hierarchical testing and validation framework on an actual testing vehicle and utilize a closed test track and parking area to arrange and set up our testing environment. Our primary goal is to evaluate the decision-making performance and scene understanding capabilities of a specific VLM model. This section details the modifications and setup for both the test vehicle and the testing environment, serving as a case study of the framework’s real-world application. Through controlled, repeatable trials under different testing environments with various conditions, we reveal the VLM’s aptitude for robust decision-making and scene interpretation. Moreover, by systematically altering the testing environment, we validate the consistency of the model’s capabilities across diverse, dynamic scenarios.

\subsection{Real-World Testing Vehicle Setup}

The proposed hierarchical testing framework for VLM-integrated autonomous driving is deployed on a hybrid vehicle with by-wire control modification. Additionally, we deployed and integrated the Autoware software stack as our Autonomous Driving Layer due to its open-source flexibility and ROS-based architecture. An overview of the real-world architecture of the proposed framework is shown below in Fig. \ref{fig:case}.

\subsubsection{Strategic Driving Intelligence Layer} 
In our real-world deployment, we adopt on-cloud ChatGPT-4 \cite{openai2025chatgpt} as our VLM agent in the Strategic Driving Intelligence Layer. The on-cloud implementation allows for scalable computational resources while still enabling real-time decision-making on the vehicle. In the VLM’s scene understanding and decision-making pipeline, the Prompt $\boldsymbol{P}$ will provide necessary multimodal messages in a unified structure with VLM. These messages consist of essential driving information such as a high-resolution image or video feed capturing the driving scene, including road markings, pedestrians, and other vehicles under varied environmental conditions, along with contextual natural language instructions to guide the VLM and supply temporal sequences for the tasks. This integrated multimodal prompt enables the VLM to accurately interpret complex driving environments and make informed output decisions tailored to the driving scenario. These decisions, in this case, are essentially to select appropriate driving behavior, including steering adjustments to safely navigate turns, performing lane changes when obstacles are detected, and speed control actions such as acceleration, deceleration, or emergency braking based on traffic conditions and hazards. Consequently, the system can adapt to dynamic traffic situations and maintain both safety and efficiency.



\subsubsection{Prompt Generation \& Action Layer}



The Prompt Generation and Action Layer serves as a critical intermediate processing stage that bridges the Strategic Driving Intelligence Layer with the underlying Autonomous Driving Layer. This layer consists of two primary components: the \textbf{Prompt Generation Interface} and the \textbf{Action Interface}, which work together to translate high-level driving strategies into executable vehicle behaviors.

\textbf{The Prompt Generation Interface} implements a structured query-response mechanism that processes the input from the Strategic Driving Intelligence Layer and generates appropriate prompts with the unified format for the VLM agent. In this case, as our testing goal is to test and validate the abilities of the VLM in decision-making and scene understanding, the Prompt $\boldsymbol{P}$ needs to contain a multimodal input that closely mirrors the real-world data an autonomous vehicle would process. These inputs include various visual data that capture complex driving scenes and feature multiple elements, the contextual and instructional text of descriptions or queries that specify the task at hand, and temporal information about vehicle status. Thus, we divide these contents into three main parts:
     \begin{itemize}
        \item a) System Statement: These messages describe the task at hand and provide high-level driving logic for the entire system using chain-of-thought prompting. The system messages reinforce the test’s objective of validating the agent’s decision-making consistency and scene understanding, which guide the VLM agent's reasoning process and ensure that the generated actions align with the overall driving objectives and constraints.
    \end{itemize}
        \begin{align}
        \boldsymbol{System Statement} &\to \label{eq:system_messages}\\
        &\left\{\begin{array}{l}
        {\footnotesize\text{You are a fully autonomous vehicle operating on a public roadway.}}\\
        {\footnotesize\text{You receive real-time sensor feeds from the front-view camera}}\\
        {\footnotesize\text{and from the on-board passenger, who provides driving preferences and needs.}}\\
        {\footnotesize\text{The possible driving behaviors for the current scenario will be provided.}}\\
        {\footnotesize\text{You need to take action to choose the optimal driving behavior}}\\
        {\footnotesize \text{from all possible behaviors and refine the motion control.}}\\
        {\footnotesize \vdots}\\
        \end{array}\right. \nonumber
        \end{align}
    \begin{itemize}
        \item b) Few Shots Examples: The Prompt Generation Interface includes a set of carefully curated examples that demonstrate the desired input-output behavior for the VLM agent. These examples serve as a reference for the VLM agent, helping it understand the expected format and content of the prompts and the corresponding actions. In the sample query,  this process showcases how the model synthesizes vehicle state, available behaviors, and human preferences into a coherent plan, serving our test goal of verifying the VLM’s ability to respond to real-world scenarios.
    \end{itemize}
        \begin{align}
        \boldsymbol{E} &\to \label{eq:examples}\\
        &\left\{\begin{array}{l}
        {\footnotesize \text{\textbf{Query}: The current vehicle state is} \textcolor{red}{[130, 46, 0.97, 38, highway]}}\\
        {\footnotesize \text{The possible driving behavior is} \textcolor{red}{[yield, overtaking, following]}}\\
        {\footnotesize \text{The passenger's command is} \textcolor{red}{[driving\_more\_stable]}}\\
        {\footnotesize \text{\textbf{Thought}: The passenger wants me to drive more stably on the highway.}}\\
        {\footnotesize\text{To achieve this, I should maintain a safe following distance}}\\
        {\footnotesize \text{and avoid sudden lane changes or overtaking}}\\
        {\footnotesize \text{\textbf{Action}: Selected Driving Behavior:} \textcolor{red}{[following]}}\\
        {\footnotesize \text{Longitudinal Control Params:} \textcolor{red}{[1.1, 0.02, 0.01]}}\\
        {\footnotesize \text{Lateral Control Params:} \textcolor{red}{[0.2, 0.35, 2.0]}}\\
        {\footnotesize \vdots}
        \end{array}\right. \nonumber
        \end{align}
        \begin{itemize}
        \item c) Vehicle Status Description: The Prompt Generation Interface utilizes the information in the Autonomous Driving State Vector $\boldsymbol{V}_s$ to provide a detailed description of the vehicle's current status. The provided status information includes position, heading, velocity, possible driving behaviors, and passenger commands, as well as encoded a frame of video processed by the perception module from the front view camera. By combining these inputs, a comprehensive snapshot of the current environment is provided with the under-testing VLM. This setup allows us to rigorously evaluate how well the VLM interprets real-time sensory input and adapts its driving strategy, thereby fulfilling our testing goals of assessing scene understanding and decision-making.
        \end{itemize}
            \begin{align}
            \boldsymbol{D} &\to \label{eq:vehicle_status}\\
            &\left\{\begin{array}{l}
            \quad {\footnotesize \text{The current vehicle state is} \textcolor{red}{[x_{c}, y_{c}, \psi_{c}, v_{c}, m_{c}]}}\\
            \quad {\footnotesize\text{The front view image captured by is} \textcolor{red}{[encoded_image]}}\\
            \quad {\footnotesize \text{The possible driving behavior is} \textcolor{red}{[b_{1}, b_{2}, ..., b_{n}]}}\\
            \quad {\footnotesize\text{The passenger's command is} \textcolor{red}{[command]}}\\
            \end{array}\right. \nonumber
            \end{align}
        
\textbf{The Action Interface} receives the VLM agent's output decision and processes it to translate a set of actionable commands tailored to the driving scenario by generating executable vehicle behaviors.  In the VLM decision-making case, this decision can be concentrated into three types, including steering adjustments, appropriate driving behaviors, and speed control actions. For instance, based on the provided examples and the VLM agent's reasoning, the Action Interface selects the most appropriate trajectory that aligns with the current transportation scenario. It then adjusts the lateral and longitudinal controller parameters to achieve the desired performance. The detailed action is shown as follows:
\begin{align}
    \boldsymbol{A} &\to \label{eq:action_interface}\\
    &\left\{\begin{array}{l}
    \quad {\footnotesize
      \text{Selected Driving Behavior: }
      \textcolor{red}{{\footnotesize [Yield]}}
    }\\
    \quad {\footnotesize
      \text{Lateral Control Params: }
      \textcolor{red}{{\footnotesize [p_{1}, p_{2}, p_{3}]}}
    }\\
    \quad {\footnotesize
      \text{Longitudinal Control Params: }
      \textcolor{red}{{\footnotesize [p_{4}, p_{5}, p_{6}]}}
    }\\
    \end{array}\right. \nonumber
\end{align}

\subsubsection{Information Process \& Execution Layer}

In the Information Process Phase, multiple specialized processors in this layer continuously extract and concentrate real-time high-relevance messages required by the Prompt Interface to an Autonomous Driving State Vector $\boldsymbol{V_s}$ so that the irrelevant message can be filtered to increase the information transmission efficiency. As the testing goal requirement in this case, this vector should contain perception results, safety-guaranteed trajectories from the planning module, real-time vehicle states, and human feedback verbal command.

Simultaneously, to achieve the VLM's requests for desired driving behavior, steering, and speed adjustment, processors in the Execution Phase refine overall driving performance by implementing the planned trajectory that best aligns with the most recently optimized driving strategy and by fine-tuning the motion control parameters of the ego vehicle. These two phases work separately to provide rich information, allowing the VLM to provide flexible, precise decisions while the established Autonomous Driving functionality stack ensures the desired maneuvering at the vehicle level by following the decision.

\textbf{Information Process Phase} In the Information Processing Phase, dedicated processors are applied to capture multiple pivotal messages from various autonomous driving function modules for the VLM agent reasoning and processing.
\begin{itemize}
    \item \textbf{Perception Results Aggregator} In our special testing of VLM decision-making ability, perception feed $\boldsymbol{F}$ is a camera image bundled together with object detection results of YOLO \cite{su141912274}, also along with detected objects textual information and bounding boxes. This perception feed $\boldsymbol{F}$ will be integrated into the autonomous driving state Vector $\boldsymbol{V_s}$ for the following process.
    \item \textbf{Planning and Navigation Acquisition} In order to let the VLM make the best decision, comprehensive planning \& navigation results are essential. This module obtains all possible safety-assured driving behaviors, represented as $\boldsymbol{(b_1, b_2, ..., b_n) \in B}$. In our case, as the planning results, multiple feasible way-point-based trajectories $\boldsymbol{b_i}$ will be provided for adjustment and selection to achieve the best aliment with the VLM desired maneuvers. 
    \item \textbf{Localization State Adapter} This module acquire vehicle states $\boldsymbol{S (x,y,\psi, v, m)}$, where $\boldsymbol{x}$, $\boldsymbol{y}$, $\boldsymbol{\psi}$, and $\boldsymbol{v}$ represent the global position, orientation, and velocity of the ego vehicle by subscribing relevant topics in the Autonomous Driving Layer. $\boldsymbol{m}$ represents the map information. Besides, the map information will also be provided as a string vector, which potentially helps the VLM understand the scene. 
    \item \textbf{Speech Command Processor} To achieve personalized autonomous driving performance by interacting with human verbal commands, a microphone continuously records the acoustic signal inside the cabinet and sends it to the Speech Command Processor module for further analysis. To capture human verbal command $\boldsymbol{H}$ precisely in real-time, we utilize a state-of-the-art voice recognition technology, the open-source API Whisper \cite{chidhambararajan2022efficientword}. Once verbal commands are accurately captured, the Speech Command Processor accordingly translates these verbal commands into a string vector through ROS topic. This translation is crucial for ensuring that the contents and specificities of the human’s spoken words are effectively converted into prompts for the following VLM.
\end{itemize}

\textbf{Excuation Phase} In the Execution Phase, the high-level actions generated by the VLM agent are translated into practical steps through executable shell scripts. The execution processors in this layer are responsible for converting the action commands into vehicle-specific control inputs while ensuring safety constraints are met.

The updated driving behavior $\boldsymbol{b_i \in B}$ and the tuned control parameters are sent to the Autonomous Vehicle Layer, where they are executed by the vehicle's actuators using standard control algorithms. The Execution Phase consists of two core processors:
\begin{itemize}
    \item \textbf{Driving Behavior Selection}: The Driving Behavior Selection processor receives the selected trajectory from the VLM agent and processes it to ensure compatibility with the vehicle's control system. This module may perform additional checks, such as verifying the feasibility of the trajectory given the vehicle's current state and the surrounding environment. Once validated, the selected trajectory is passed on to the Motion Control Adjustment module.
    
    \item \textbf{Motion Control Adjustment}: The Motion Control Adjustment processor is responsible for updating the vehicle's control parameters based on the VLM agent's output. This is achieved by modifying the relevant ROS parameters, which are then used by the vehicle's control algorithms. The module ensures that the updated control parameters remain within a safety-guaranteed range to maintain vehicle stability and passenger comfort.
\end{itemize}

The execution of this processor is realized through shell scripts, which automate the process of updating ROS parameters and triggering the appropriate control algorithms. For example, consider the following shell commands:

\begin{minipage}{\linewidth}

\begin{flalign}
p \to \quad && 
\end{flalign}

\begin{lstlisting}[language=bash]
ros param set @\textcolor{red}{/kp 1.1 /ki 0.02 /ki 0.01}@
ros param set @\textcolor{red}{/weight\_lateral\_err 0.2 /weight\_heading\_err 0.35 /speed\_steering\_input\_coefficient 2.0}@
./scripts/yield.sh
...
\end{lstlisting}

\end{minipage}

In this example, the first command sets the PID controller gains for the vehicle's 
longitudinal control system. The second command adjusts the weights for the lateral error, heading error, and speed-steering input coefficient, which the Model Predictive Controller uses for the vehicle's Lateral control. Finally, the last command executes a shell script named yield.sh, which contains a sequence of instructions to perform the yielding maneuver. By including the execution logic within shell scripts, the Execution Phase allows for easy adaptation to different vehicle platforms and control architectures while maintaining a clear separation between the decision-making and execution processes.

\subsubsection{Autonomous Driving Functionality Layer}

We implement Autoware.AI as our primary autonomous driving functionality backbone due to its open-source flexibility, comprehensive software stack, and robust ROS-based architecture. Autoware.AI provides essential functionalities, enabling seamless customization and integration of various algorithms to meet specific project requirements. Its modular design supports independent development and testing of components, enhancing scalability across different hardware configurations. 

In our vehicle's implementation, we deploy YOLOv5 as the primary perception module for object detection and recognition. The object detection process can be formulated as follows:
\begin{equation}
\boldsymbol{b}, \boldsymbol{c}, \boldsymbol{p} = f_{YOLO}(\boldsymbol{I})
\end{equation}
Where $\boldsymbol{I}$ is the input image, $f_{YOLO}$ represents the YOLOv5 model, and $\boldsymbol{b}$, $\boldsymbol{c}$, and $\boldsymbol{p}$ are the bounding boxes, class labels, and confidence scores of the detected objects, respectively. These perception results are potentially processed and integrated into the Prompt $\boldsymbol{P}$ by following provided with VLMs for a better understanding of the perception results.

For localization, we apply the Normal Distributions Transform (NDT) matching algorithm \cite{1249285}, which estimates the vehicle's pose by aligning the current LiDAR scan with a pre-built map. The NDT algorithm minimizes an objective function that reflects the distribution of points, aiming to find the optimal position where the scan best aligns with the map. To further enhance localization accuracy and robustness, additional sensors, including GNSS and IMU, are integrated to provide precise measurements of the vehicle’s position, velocity, acceleration, and orientation. Combined with detailed map information, these inputs offer a comprehensive description of the vehicle state, supporting accurate scene interpretation and decision-making by the VLM.

In our testing case study, multiple pre-recorded trajectories are used to represent the planning results for different driving behaviors, with each trajectory consisting of a sequence of waypoints. This pre-recorded trajectory representation ensures consistency in the planned paths, allowing the vehicle to follow the same trajectory across repeated trials. By doing so, we maintain strict control over experimental variables, enabling fair comparisons and accurate validation of the VLM’s decision-making and scene understanding performance under identical conditions.

For vehicle motion control, we adopt a decoupled control strategy that uses a PID controller for longitudinal control and Model Predictive Control (MPC) for lateral control. The PID controller adjusts the throttle and brake commands based on the error between the desired and actual vehicle speed. The control law for the PID controller can be expressed as:
\begin{equation}
u[k] = K_p e[k] + K_i \sum_{i=0}^{k} e[i] \Delta t + K_d \frac{e[k] - e[k-1]}{\Delta t}
\end{equation}
Where $\boldsymbol{u(t)}$ is the control signal, $\boldsymbol{e(t)}$ is the error signal, $\boldsymbol{\Delta t}$ is the sampling interval, and $\boldsymbol{K_p}$, $\boldsymbol{K_i}$, and $\boldsymbol{K_d}$ are the proportional, integral, and derivative gains, respectively. By tuning these gains, the vehicle can achieve different driving styles with quick or slow responses.

The MPC controller for lateral control optimizes the steering angle by minimizing a cost function that considers the lateral tracking error, heading error, and control effort. The MPC optimization problem can be formulated as:
\begin{equation}
\begin{aligned}
\min_{\mathbf{u}} \quad 
& \sum_{k=0}^{N-1} \bigl(Q^T\|\mathbf{y}_k - r\|\;Q \;+\; \|\mathbf{u}_k\|\;R^2\bigr)
  \;+\;\|\mathbf{y}_N - r\|\;P^2 \\
\text{s.t.}\quad 
& \mathbf{x}_{k+1} \;=\; f\bigl(\mathbf{x}_k,\;\mathbf{u}_k\bigr),
  \quad k = 0,\dots,N-1,\\[6pt]
& \mathbf{y}_k \;=\; g\bigl(\mathbf{x}_k\bigr),
  \quad k = 0,\dots,N,\\[6pt]
& \mathbf{u}_{\min}\;\le\;\mathbf{u}_k\;\le\;\mathbf{u}_{\max},
  \quad k = 0,\dots,N-1.
\end{aligned}
\end{equation}
Where $\boldsymbol{x}_k$, $\boldsymbol{u}_k$, and $\boldsymbol{y}_k$ are the state, control input, and output at time step $k$, respectively, $\boldsymbol{r}_k$ is the reference trajectory, $\boldsymbol{Q}$, $\boldsymbol{R}$, and $\boldsymbol{P}$ are weighting matrices, and $N$ is the prediction horizon. The functions $\boldsymbol{f(\cdot)}$ and $\boldsymbol{g(\cdot)}$ represent the system dynamics and output equations, respectively. By adjusting the weights in the cost function, the MPC controller can adapt the vehicle's driving style, such as minimizing lateral jerk for a smoother ride.

The integration of these advanced algorithms within the Autoware.AI framework enables our autonomous vehicle to perceive its surroundings accurately, localize itself within the environment, plan safe and efficient trajectories, and execute precise control commands to achieve the desired driving behavior.

\begin{figure}[H]
  \centering
  \includegraphics[width=0.75\linewidth]{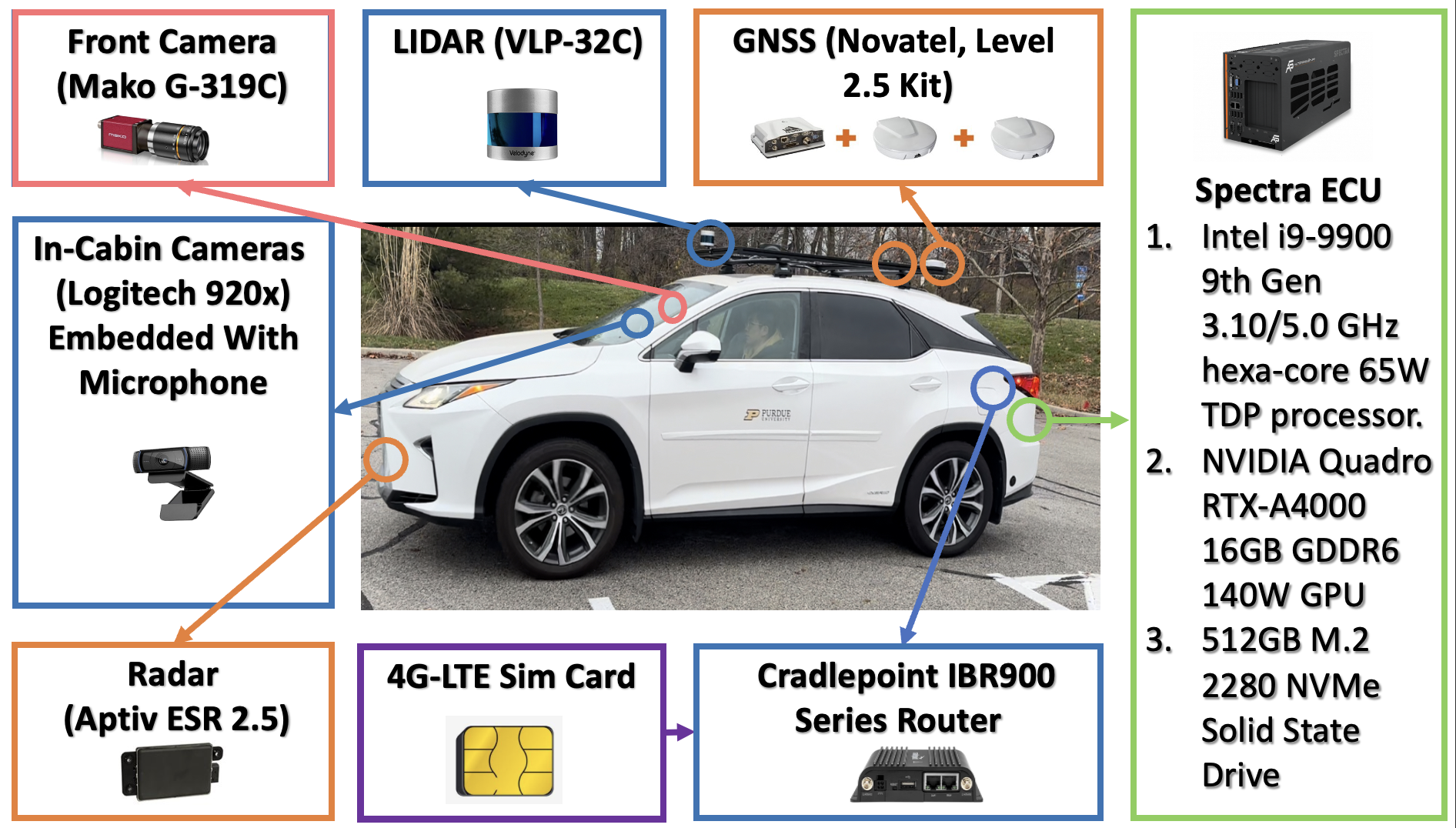}
  \caption{The Hardware Layer setup of the autonomous driving enabled vehicle that is used in our case study. This vehicle is modified by AutonomouStuff with advanced sensor kits, an on-board ECU, and a By-wire-control system. We implement our proposed hierarchical testing framework on this specific vehicle. }
  \Description{This vehicle is modified by AutonomouStuff with advanced sensor kits, on-board ECU, and By-wire-control system. We implement our proposed hierarchical testing framework on this specific vehicle}
  \label{fig:setup}
\end{figure}

\subsubsection{Hardware Layer}
The modification on the 2019 Lexus 450h is conducted by AutonomouStuff \cite{autonomoustuff}; this autonomous driving vehicle is equipped with comprehensive subsystems, including a power distribution system, connection router, sensors suite, PACMod drive-by-wire system, and a high-performance on-board Electronic Control Unit (ECU). An overview of the vehicle hardware setup is shown in Fig.~\ref{fig:setup}.

\textbf{Sensors configuration and management} The sensor suite in the vehicle includes a Velodyne VLP-32C LiDAR, a radar (Aptiv ESR 2.5 24V), two front-view cameras (Mako G-319C), and a real-time kinematic (RTK) correction enabled GNSS Positioning Kit with an inertial measurement unit kit (Novatel Level 2.5 kit). Specifically, each of these sensors is able to provide raw data at 10 fps through an Ethernet port connection. In addition, two webcams are mounted near the front seat and on the windshield to capture driver behavior, where the built-in microphones can simultaneously capture verbal commands from drivers or passengers. 




\textbf{Connectivity \& Computing} A Cradlepoint IBR900 Series Router embedded with a 4G-LTE AT\&T SIM card is installed to provide a cellular connection for the vehicle. It offers a ping latency of 30-50 ms and bandwidths of 50-70 Mbps for uploads and 50-60 Mbps for downloads.

A high-performance ECU is installed in the vehicle's back trunk. The core computational configuration is below: Intel i9-9900 9th Gen 3.10/5.0GHz Hexa-core 65W processor with eight cores and 16 threads, 64GB RAM, NVIDIA Quadro RTX-A4000 16GB GPU, and 512GB NVMe solid-state drive.

\textbf{By-Wire Control}
The on-board By-Wire control \cite{Song01022015} is PACMod 3.0 System provided by AutonomouStuff \cite{autonomoustuff}. This Drive-by-Wire system consists of two main subsystems: Speed and Steering Control (SSC) \cite{SSC} and PACMod system. SSC is a software package that assists the autonomous driving stack in integrating with the Drive-by-Wire system by unifying control message types. This allows the PACMod system to use this control signal to actually access the CAN bus and control the accelerator, brake, shifter, and turning signals. 

\subsection{Real-World Testing Experiment Setup}
\begin{figure}[ht]
  \centering
  \includegraphics[width=\linewidth]{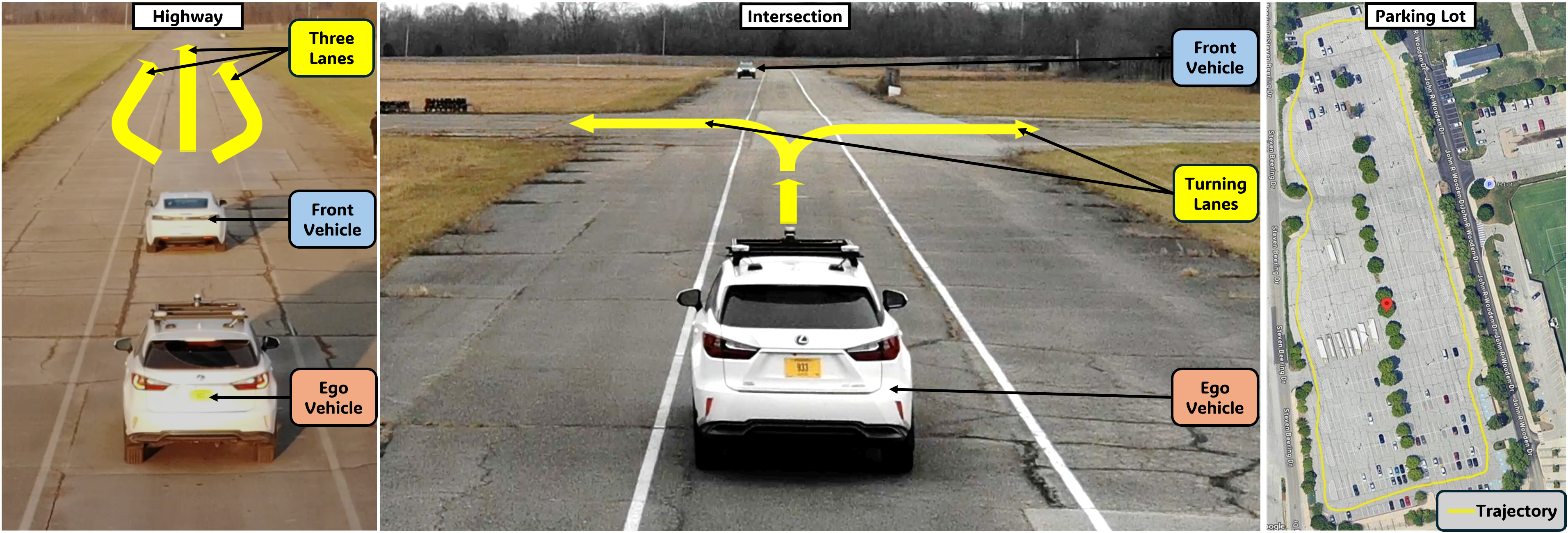}
  \caption{Illustration of the real-world test environments used in our case study. The closed test track (left and center images) accommodates both highway and intersection scenarios, allowing controlled evaluation of lane changes, following behavior, and turning maneuvers. The private parking lot (right image) is used to test low-speed navigation and parking tasks. These diverse, real-world settings enable a comprehensive assessment of our VLM-based autonomous driving framework under varying road and traffic conditions with diverse traffic participants.}
  \Description{}
  \label{fig:track}
\end{figure}

To evaluate the performance of our autonomous vehicle in realistic driving scenarios, we design and deploy three independent testing scenarios to conduct experiments: a three-lane highway, a two-lane intersection, and a parking lot track, as shown in Fig. \ref{fig:track}. The highway and intersection tests are carried out at a dedicated test track located at 950 S 450 W, Columbus, IN, USA. This controlled environment allowed for the safe testing of high-speed maneuvers and complex traffic interactions. The parking lot track experiments are conducted at the North Stadium Parking Lot in West Lafayette, IN, USA, which provides a more confined space to assess the vehicle's low-speed maneuvering capabilities with more uncertainty.

We design a series of test scenarios for each environment, covering common driving situations. These test cases included:
\begin{itemize}
\item Highway: Maintaining a constant speed, performing lane changes, reacting to sudden slowdowns in traffic, and merging onto the highway from an on-ramp. 
\item Intersection: Navigating four-way stops, yielding to pedestrians, making protected and unprotected left turns, and responding to traffic signals.
\item Parking lot: Navigating narrow lanes, avoiding static and dynamic obstacles, parallel parking, and backing into a parking spot. 
\end{itemize}
 
For each trip in our experiment, we call the VLM on the cloud multiple times and record the computational resources, times, and messages for validation and assessment, as discussed in Section \ref{sec:Testing and Validation}.
Once the experiments start, we send the prompting request to the on-cloud VLM with a frequency of 3 seconds per request to avoid latency and message loss due to VLM reasoning. As mentioned, each prompt contains multiple outputs from the lower Autonomous Driving Layer: the vision results feed from the perception module, vehicle current states from the localization module, feasible trajectories from the planning module, and detected verbal commands from passengers. These messages among the middle layer of our proposed framework are updated with a faster frequency, with a minimum of 10 HZ. Notably, the speech command processor on our vehicle will update once a new command has been detected and processed. Otherwise, it will keep publishing the previous command as human input. Once the action is generated by the VLM agent on the top layer, the results will be executed by finalizing the planning results and adjusting the motion control parameters.

 \begin{figure}[H]
   \centering
   \includegraphics[width=\linewidth]{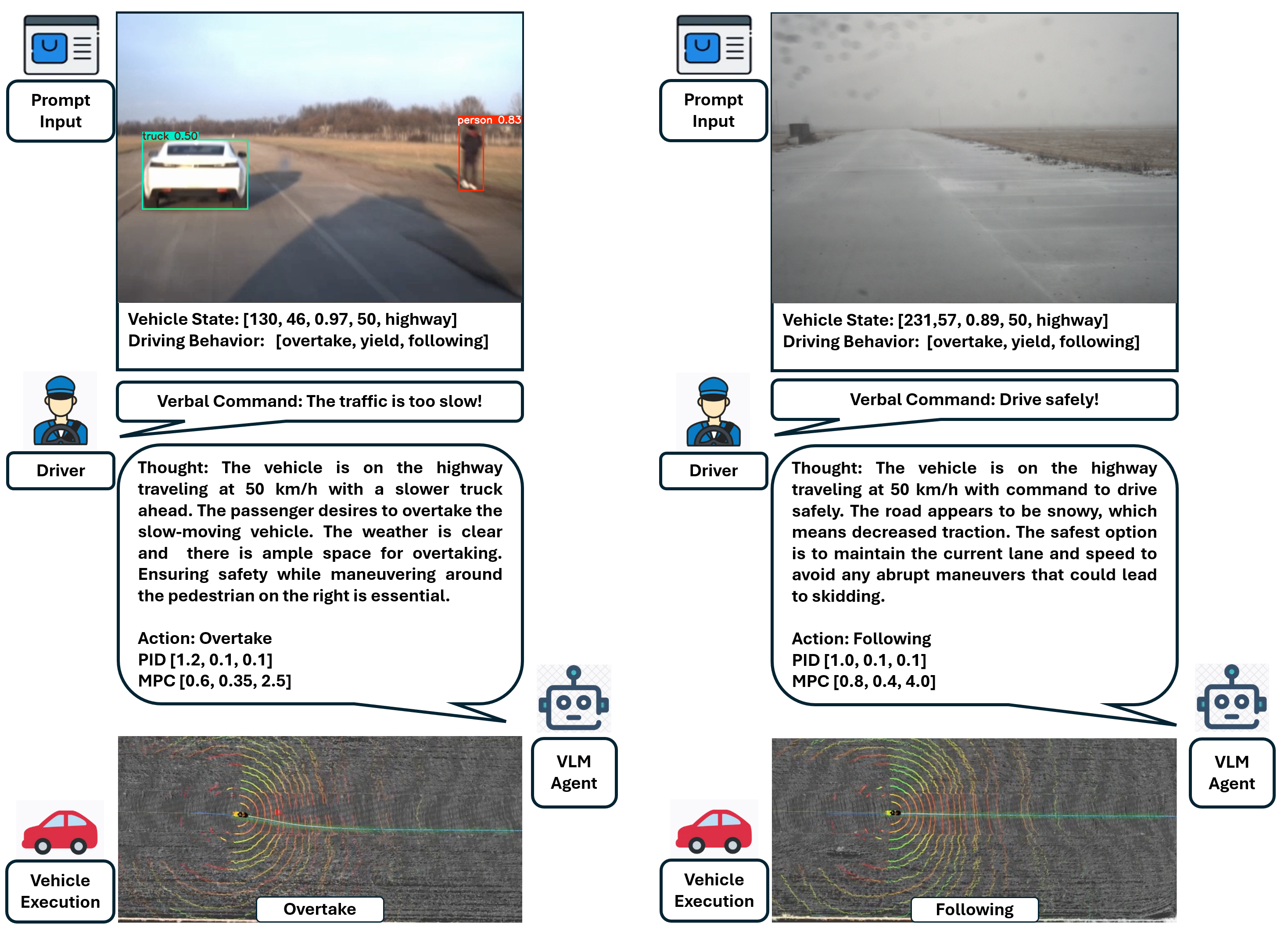}
   \caption{Illustration of the highway scenario demonstrating two distinct driving conditions for our testing platform. In the first condition, the vehicle travels at 50 km/h behind a slower vehicle on a clear day; the passenger’s command ``The traffic is too slow” prompts the VLM agent to generate an overtaking maneuver by adjusting motion control parameters. In the second condition, the vehicle drives on a snowy highway with low visibility and receives the command ``Drive safely." Given the adverse weather, the VLM agent reduces the proportional gain for longitudinal control to avoid sudden accelerations or braking and increases the steering weight to maintain stability, ensuring safe operation despite limited traction. }
   \Description{}
   \label{fig:highway_scenario}
 \end{figure}

\subsubsection{Highway}
\label{appendix:highway}

The highway scenario, as shown in Figure \ref{fig:highway_scenario}, presents two different driving conditions to test the autonomous vehicle's performance in maintaining a constant speed, performing lane changes, reacting to sudden slowdowns in traffic, and merging onto the highway from an on-ramp.

On the first trip, the vehicle is traveling at 50 km/h on a clear day, with a slower vehicle ahead. The passenger commands the ego vehicle to overtake the slow-moving vehicle. The prompt input to the VLM agent includes the vehicle state $\boldsymbol{S = [130, 46, 0.97, 50, highway]}$, driving behavior $ \boldsymbol{B = [overtake, yield, following]}$, and the verbal command ``The traffic is too slow." The VLM agent processes this information and generates an action to overtake the truck while adjusting the motion control parameters (PID and MPC) to ensure an appropriate maneuver.

\begin{figure}[H]
  \centering
  \includegraphics[width=\linewidth]{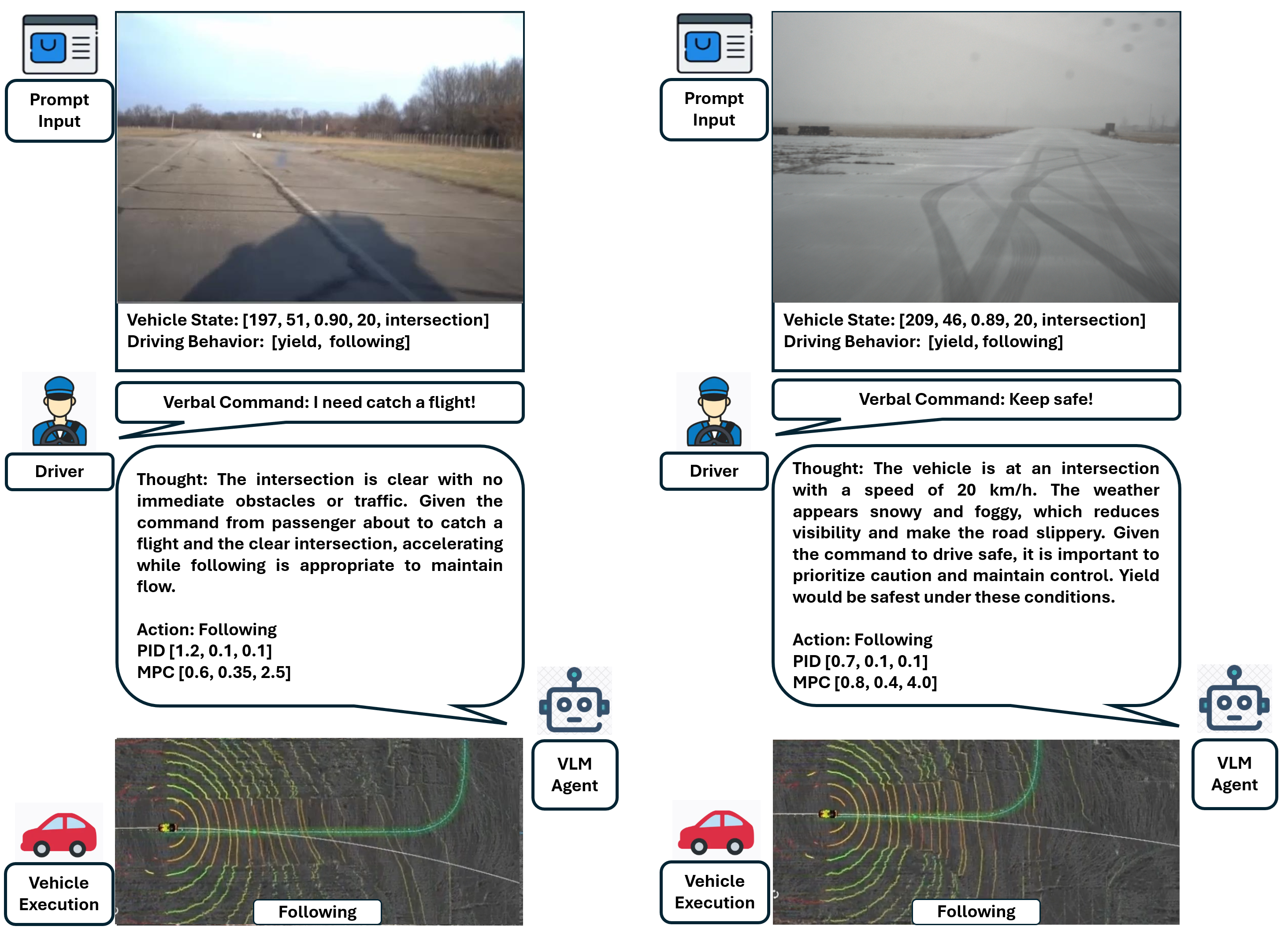}
  \caption{Illustration of the intersection scenario under two distinct environmental conditions for the proposed testing platform. With clear weather and no immediate obstacles, the urgent verbal command ''I need to catch a flight” prompts the VLM agent to accelerate while maintaining safe following distances. Under snowy, foggy conditions, the passenger requests caution ``Keep safe,” leading the VLM agent to reduce speed and adjust motion control parameters to maintain stability and avoid sudden maneuvers.}
  \Description{}
  \label{fig:intersection}
\end{figure}
\subsubsection{Intersection}
\label{appendix:Intersection}

On the second trip, the vehicle is driving on a snowy highway with low visibility due to a blurred windshield. The passenger commands the vehicle to drive safely. The prompt input to the VLM agent includes the updated vehicle state $\boldsymbol{S = [231, 57, 0.89, 20, highway]}$, driving behavior $\boldsymbol{B = [overtake, yield, following]}$, and the verbal command ``Drive safely." Given the adverse weather conditions and reduced traction, the VLM agent prioritizes caution and generates an action to maintain the current lane and speed while keeping a safe distance from the leading vehicle to avoid skidding or losing control. Specifically, for adjusting the motion control parameters, the VLM agent decreases the proportional item $\boldsymbol{K_p}$ for the longitudinal PID control to reduce the throttle and brake response speed to avoid sudden change and slip. Also, it increases the steering weight input $\boldsymbol{W_s}$ to maintain stability by limiting the steering input.

The intersection scenario, as shown in Figure \ref{fig:intersection}, tests the autonomous vehicle's ability to navigate cross traffic, yield to pedestrians, make protected and unprotected turns, and respond to traffic quickly under different environmental conditions.
\begin{figure}[H]
  \centering
  \includegraphics[width=\linewidth]{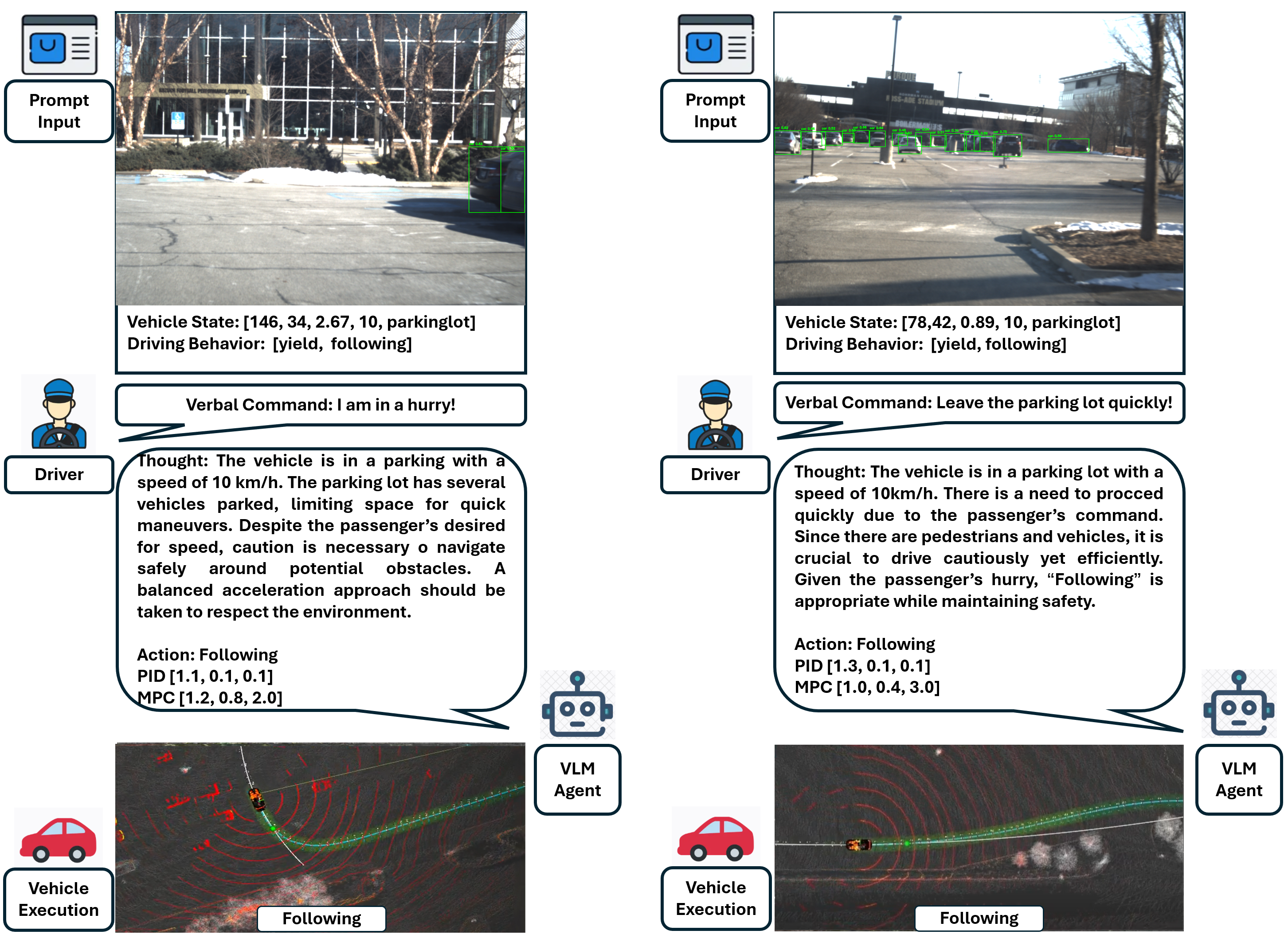}
  \caption{Illustration of the parking lot scenario under two different commands and conditions for our VLM-based framework. The vehicle travels at $10$ km/h with limited space due to parked cars. However, the passenger is in a hurry. The VLM agent balances speed and safety by navigating precisely around obstacles. In a similar $10$ km/h setting, the passenger commands the vehicle to ``Leave the parking lot quickly” the VLM agent again prioritizes safe yet efficient maneuvers by ensuring an accurate path following among the pedestrians and closely parked vehicles.}
  \Description{}
  \label{fig:parkinglot}
\end{figure}
On the first trip, the vehicle approaches a clear intersection with no immediate obstacles or traffic. The passenger provides a verbal command, ``I need to catch a flight," indicating urgency. The prompt input to the VLM agent includes the vehicle state $\boldsymbol{S =[197, 51, 0.90, 20, intersection]}$, driving behavior $\boldsymbol{B = [yield, following]}$, and the verbal command. The VLM agent processes this information and generates an action to follow the current path while accelerating appropriately to maintain traffic flow, considering potential safety measures. 

On the second trip, the vehicle approaches an intersection under snowy and foggy conditions, which reduces visibility and traction. The passenger commands the vehicle to ``Keep safe." The prompt input to the VLM agent includes the updated vehicle state $\boldsymbol{S= [209, 46, 0.89, 20, intersection]}$, driving behavior $\boldsymbol{B = [yield, following]}$, and the verbal command. Given the adverse weather conditions, the VLM agent prioritizes caution and generates an action to maintain a safe speed and following distance by adjusting PID and MPC parameters to avoid sudden maneuvers that could lead to loss of control. In this circumstance, the generated action adjusts the proportional item $\boldsymbol{K_p}$ and the input of the steering weight $\boldsymbol{W_s}$ to implement a smooth maneuver to keep the vehicle stable.

\subsubsection{Parking Lot}
\label{appendix:parkinglot}
The parking lot scenario evaluated the vehicle's ability to navigate in tight spaces, avoid obstacles, and accurately park in designated spots. Fig. \ref{fig:parkinglot} shows two examples of a vehicle navigating a parking lot under different commands and conditions.

In the first trip in the parking lot situation, the vehicle is moving at $10$ km/h, and the passenger is in a hurry. The parking lot has several vehicles parked, limiting the space for quick maneuvers. Despite the desire of the passenger for speed, the VLM agent recognizes the need for caution and generates an action to follow the path while maintaining a relatively high precision for path tracking to respect the environment and navigate safely around potential obstacles. Thus, a high weight value for the lateral tracking error $\boldsymbol{W_l}$ in the motion control parameters is implemented.

On the second trip, the prompt input includes the updated vehicle state $\boldsymbol{S = [78, 42, 0.89, 10, parkinglot]}$, driving behavior $\boldsymbol{B = [yield, following]}$, and the similar verbal command. Given the much more free space, it is crucial to drive cautiously yet efficiently. Under this situation, the VLM agent also adjusts the motion control parameters with a slightly lower lateral error weight $\boldsymbol{W_l}$ to provide less precise path tracking performance with larger acceleration by increasing the proportion gain $\boldsymbol{K_p}$.

\section{Testing and Validation}
\label{sec:Testing and Validation}

\subsection{Evaluation Metrics}

To comprehensively evaluate the performance of our proposed framework, we employ three key metrics: latency, correctness, and computational resource utilization. These metrics provide a holistic view of the system's real-time capabilities, accuracy, and efficiency by analyzing the inputs, outputs, and performance of some core modules. As in Table. \ref{tab:core_modules} shows the modules we test and their inputs and outputs.

\begin{table}[htbp]
\centering
\caption{Tested Modules and Their Inputs and Outputs in The Proposed Autonomous Driving Testing Platform}
\label{tab:core_modules}

\renewcommand{\arraystretch}{1.3}
\small

\resizebox{\linewidth}{!}{%
\begin{tabular}{p{6cm} p{8cm} p{6cm}}
\toprule
\textbf{Module} & \textbf{Key Inputs} & \textbf{Key Outputs} \\
\midrule

Prompt Generation Interface 
  & Current Vehicle State, Recognized Text Command
  & Formatted Prompt Text \\

Action Interface
  & VLM’s Selected Behavior, Desired Parameters
  & Final Trajectories \\

Speech Command Processor
  & Microphone Audio Stream
  & Recognized Command String \\

Vision Perception Aggregator
  & Image Feeds
  & Perception Results \\

Localization State Adapter
  & Localization Results
  & Vehicle States \\

Planning \& Navigation Acquisition
  & Planning Results
  & Safety Behavior \\

Driving Behavior Selection
  & VLM Recommendation
  & Confirmed Final Maneuver \\

Motion Control Refinement
  & Desired Control Parameters
  & Control Parameters Augments \\
\bottomrule
\end{tabular}%
}
\end{table}

\textbf{Time Effeciency}
Latency is a critical metric for assessing the real-time performance of an autonomous driving system. It measures the time taken by each module to process its inputs and generate outputs. In our experiment, we derive the latency $L_i$ for each module by calculating the time difference between receiving a message and presenting the information to the next layer.


Specifically for the latency of the speech command processor module, we use the processing time for each word as the latency. Due to the varying lengths of the commands, the processing time for commands varies from $0.5 s$ to $1.9 s$. 

\textbf{Accuracy}
Accuracy evaluates the correctness of the outputs generated by each module. We measure accuracy by comparing the generated outputs with the expected outputs based on the input messages. 

\textbf{Computational Resource Utilization}
Computational resource utilization measures the CPU, GPU, and memory usage of each module. This metric helps identify potential bottlenecks and optimize the system for efficient resource allocation. We define computational resource utilization as:
\begin{equation}
R_{i,cpu} = \frac{T_{i,cpu}}{T_{total,cpu}} \times 100%
\end{equation}
\begin{equation}
R_{i,gpu} = \frac{T_{i,gpu}}{T_{total,gpu}} \times 100%
\end{equation}
\begin{equation}
R_{i,mem} = \frac{M_{i,used}}{M_{total}} \times 100%
\end{equation}
where $R_{i,cpu}$, $R_{i,gpu}$, and $R_{i, mem}$ are the CPU, GPU, and memory utilization of module $i$, respectively. $T_{i,cpu}$ and $T_{i,gpu}$ are the CPU and GPU time consumed by module $i$, $T_{total,cpu}$ and $T_{total,gpu}$ are the total CPU and GPU time available, $M_{i,used}$ is the memory used by module i, and $M_{total}$ is the total memory available.

\subsection{Validation Results}

As Table \ref{tab:layers} shows, all modules in our proposed framework demonstrate high performance, low latency, and efficient computational resource utilization.

\textbf{Accuracy:}
Most modules achieve $100\%$ correctness, indicating that they consistently produce accurate outputs based on their input data. The Speech Command Processor shows a slightly lower accuracy rate of $91.78\%$, which can be attributed to noise in the audio stream. This is also related to the different speech recognition tools that are used. Despite this minor shortfall, the overall system maintains a high level of accuracy across all modules.

\textbf{Latency:}
The latency measurements reveal that all modules have short processing times, enabling real-time information processing for the VLM and ensuring prompt responses to dynamic driving situations. Specifically, the Action Interface has an extremely low average latency of $2.48 \times 10^{-3} ms$, and the Prompt Generation Interface also demonstrates efficient performance with an average latency of $5.84 ms$ with a maximum latency of less than $20 ms$, which has similar performance to the Vision Perception Aggregator. Even though the Prompt Generation Interface and Vision Perception Aggregator modules have a relatively high average latency because of handling larger image data volumes that slightly elevate its average latency, these low latencies ensure that the VLM's decisions are communicated to the Execution Layer with minimal delay, thus supporting seamless integration into the autonomous driving pipeline. Notably, the Speech Command Processor exhibits a higher overall processing time, around $167.95 ms $, when considering the full end-to-end speech recognition process during the actual real-world experiments. This increase is primarily due to overhead from PyAudio and the speech recognition libraries. However, the observed latency remains within acceptable bounds for real-time operation. The overall latency is low enough to ensure that the Vision-Language Model (VLM)’s decisions are communicated to the Execution Layer with minimal delay, thus supporting seamless integration into the autonomous driving pipeline.

\begin{table}[h]
\centering
\caption{System Components Latency and Performance}
\label{tab:layers}
\resizebox{\linewidth}{!}{%
\begin{tabular}{lcccccc}
\toprule
\textbf{Module} & \textbf{Average Latency (ms)} & \textbf{Latency Standard Deviation (ms)} & \textbf{Max Latency (ms)} & \textbf{CPU (\%)} & \textbf{Memory (\%)} & \textbf{GPU (\%)} \\
\midrule
Prompt Generation Interface       & 5.84               & 0.59    & 17.77    & <1   & <1   & <1 \\
Action Interface                  & $2.48 \times 10^{-3}$ & $0.25 \times 10^{-3}$ & $29.24 \times 10^{-3}$ & <1   & <1   & <1 \\
Speech Command Processor          & 0.24       & 0.04 & 3.12 & <2   & <2   & <1 \\
Vision Perception Aggregator      & 5.71               & 0.58    & 11.76    & <3   & <5   & <1 \\
Localization State Adapter        & 0.68               & 0.14    & 4.75     & <1   & <1   & <1 \\
Planning \& Navigation Acquisition & 0.59               & 0.11    & 7.23     & <1   & <1   & <1 \\
Driving Behavior Selection        & 0.69               & 0.15    & 5.51     & <1   & <1   & <1 \\
Motion Control Refinement         & 0.88               & 0.15    & 6.52     & <1   & <1   & <1 \\
\bottomrule
\end{tabular}%
}
\end{table}

\textbf{Computational Resource}
The computational resource utilization metrics demonstrate that all modules efficiently use the available CPU, memory, and GPU resources. The Vision Perception Aggregator has the highest resource utilization, occupying $3\%$ CPU, $5\%$ of memory, and $1\%$ GPU memory, which is expected given the computationally intensive nature of processing image feeds. However, this utilization does not significantly impact the performance of the VLM or the overall autonomous driving functionality, as evidenced by the low latencies and high correctness scores. The other modules have lower computational resource requirements, with most utilizing less than $1\%$ CPU, 1 GB of memory, and $1\%$ GPU. 

This efficient resource allocation allows the on-board ECU to manage all tasks effectively without compromising real-time responsiveness. Moreover, minimizing computational overhead is critical for reserving sufficient capacity for both the autonomous driving functionality and the Vision-Language Models (VLMs), which often demand substantially more processing power.

In conclusion, the validation results demonstrate that our proposed framework achieves high correctness, low latency, and efficient computational resource utilization across all modules. These results highlight the effectiveness of our design in integrating VLM with autonomous driving functionality while maintaining real-time performance and accuracy.


\section{Conclusion and Future Work}
\label{sec:Conclusion and Future Work}


In this paper, we present a real-world autonomous driving test platform designed to validate and evaluate Vision Language Models (VLMs)-integrated autonomous driving technology. Our proposed hierarchical framework enables a high-level VLM agent to interact seamlessly with the classical autonomous functionality stack, thereby facilitating comprehensive validation across a range of designed driving scenarios. A detailed case study featuring a real-world deployment on an actual vehicle using the open-source Autoware software stack demonstrated the practical feasibility and effectiveness of our approach.

Extensive experiments across diverse scenarios revealed that our framework could achieve a latency of less than $20$ ms while maintaining high accuracy in information processing and operating with limited computational resources. These results not only verify the potential capability for effective validation ability for VLM-integrate autonomous driving, ranging from handling high-level semantic reasoning, decision-making, and specific planning and control tasks, but also highlight the significant potential for both academic research and industrial applications.


Nonetheless, several challenges remain. Future work should aim to optimize the system further to reduce latency and enhance accuracy, with a particular focus on improving the Human Verbal Command Processor. Moreover, exploring the integration of on-board VLM testing and end-to-end autonomous driving approaches could lead to even more efficient and robust systems. As VLM technology continues to evolve, maintaining a strong emphasis on real-world scenario testing and validation will be essential to ensure the safety, reliability, and performance of these systems in complex and dynamic driving environments.





\bibliographystyle{ACM-Reference-Format}
\bibliography{sample-base}
\end{document}